\documentclass{article}

\usepackage[dblblindworkshop, final]{neurips_2025}
\workshoptitle{Deep Learning for Code in the Agentic Era}

\usepackage[utf8]{inputenc} 
\usepackage[T1]{fontenc}    
\usepackage{hyperref}       
\usepackage{url}            
\usepackage{booktabs}       
\usepackage{amsfonts}       
\usepackage{nicefrac}       
\usepackage{microtype}      
\usepackage{xcolor}         
\usepackage[pdftex]{graphicx}
\usepackage{listings}
\title{Learning From Design Procedure To Generate CAD Programs for Data Augmentation}

\author{%
Yan-Ying Chen, Dule Shu, Matthew Hong, Andrew Taber, Jonathan Li, Matthew Klenk \\
  Toyota Research Institute
  Pittsburgh, PA 15213 \\
  \texttt{yan-ying.chen@tri.global} \\
}

\begin{document}

\maketitle

\begin{abstract}
Large Language Models (LLMs) have demonstrated impressive capabilities in a wide range of code generation tasks. However, generating code for certain domains remains challenging. One such domain is Computer-Aided Design (CAD) program, where the goal is to produce scripted parametric models that define object geometry for precise design and manufacturing applications. A key challenge in LLM-based CAD program generation is the limited geometric complexity of generated shapes compared to those found in real-world industrial designs. This shortfall is in part due to the lack of diversity in the available CAD program training data. To address this, we propose a novel data augmentation paradigm that prompts an LLM to generate CAD programs conditioned on a reference surface program and a modeling procedure - an idea inspired by practices in industrial design. By varying the reference surface using a collection of organic shapes, our method enriches the geometric distribution of generated CAD models. In particular, it introduces edges and faces defined by spline-based curvature, which are typically missing or underrepresented in existing open-source CAD program datasets. Experiments show that our method produces CAD samples with significantly greater geometric diversity and a higher resemblance to industry-grade CAD designs in terms of the proportion of organic shape primitives. This enhancement makes our CAD data augmentation approach a useful tool for training LLMs and other deep learning models in CAD generation. 
\end{abstract}

\section{Introduction}
\begin{figure}[!htbp]
    \centering
    \includegraphics[width=0.6\textwidth]{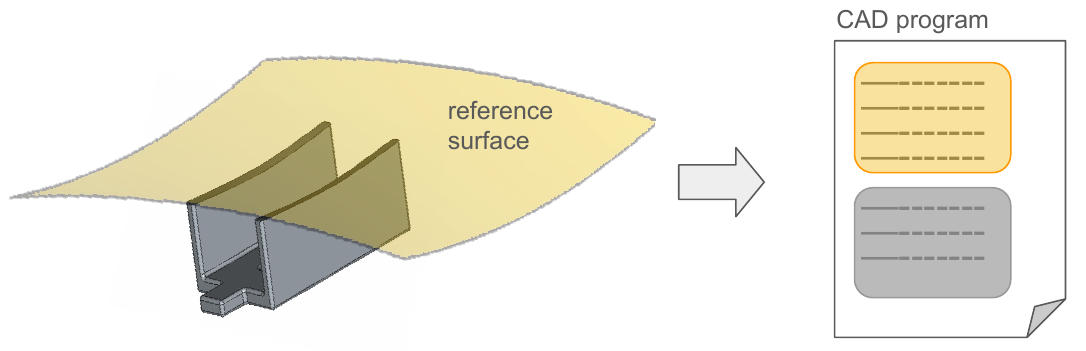}
    \caption{It is common to use a reference surface to guide CAD creation for specific design intent, e.g., compatibility to other component. Motivated by this idea, we propose to a new design procedure prompting to guide the CAD program generation toward more organic shapes.}
    \label{fig:modeling-procedure}
\end{figure}
Computer-Aided Design (CAD) is important for product design and engineering because it carries high-fidelity, parametric 3D information such as parts and topological structures that are essential for engineering and manufacturing physical products. Designers have been using CAD software, e.g., Solidworks and Fusion 360, to create CAD models through a sequence of operations such as sketch and extrude. These operations can be interpreted to different formats of CAD model representation (e.g., B-rep, boundary representation, in a STEP file) or 3D artifacts (e.g., mesh in a STL file) for a variety of downstream work such as physical simulation \cite{pfaff2021learning} and shape optimization \cite{li2022shapeoptimization}.

CAD is a classic example of visual programming, where a CAD program is created by a sequence of operations that define a 3D shape. As a vital step towards automated CAD programming, larger CAD program datasets \cite{wu2021iccv, willis2020fusion} have been proposed for training CAD program generation models. However, it is much more challenging to scale up than other visual data formats such as images or videos, because CAD programming involves multiple modalities and requires domain-specific expertise \cite{khan2024textcad} to create data. In addition, CAD data are mostly in closed systems and spread across different CAD tools without organically growing data sources similar to repositories for software programming, e.g., GitHub. Hence, the existing CAD program datasets are mostly limited in a small set of operation commands and result in simple CAD data that are far from industrial designs \cite{kimmel2025position}.

We argue that a key recipe for scaling is a sustainable eco-system to synthesize CAD programs for data augmentation. We leverage recent efforts in bridging CAD programming to generic programming languages, such as the python script libraries CadQuery\cite{au_2024_10513848} and Build123d\footnote{\url{https://github.com/gumyr/build123d}}. This exciting progress enables the application of LLM-based knowledge of common programming languages (e.g., python) to CAD programming. This advantage drives the trend of finetuning approaches \cite{rukhovich2025iccv, xu2024CADMLLM, doris2025cad} built upon those libraries and LLMs. This trend brings a turning point of architecting a data simulation and augmentation system to scale up the data needed for training CAD generation models. However, how to prompt these LLMs to simulate the CAD design process and synthesize data with geometric properties similar to industrial design standards remains an underexplored topic.

Like software engineers, CAD engineers/designers embed deep expertise and tacit knowledge in both the design processes and domains they operate in. A common strategy used by CAD designers is a modeling procedure where they create a sequence of operations from a chosen reference surface; for example, a wall that a bracket is supposed to support as illustrated the left in Figure \ref{fig:modeling-procedure}. The reference surface significantly impacts the resulting design because the sequential operations can create ripple effects, e.g., all curves adjacent to a surface must align with a part of its curvature. Hence, an internal bracket may comprise organic shapes (e.g., free form B-Spline shapes) in addition to standard primitives (e.g., flat rectangles and cylinders) because it needs to match the design's exterior. In addition, designers use field-specific design languages (e.g., double-spoke, Y-spoke, the number of spokes in a car wheel) to guide their design creation \cite{wang2024idetc}. These semantically encoded design patterns hold significant meaning for industrial design practices as they embody human-interpretable representations that play a critical role in making the generated program align more closely with design intent. The combination of reference surface, design language and procedure constitute a multimodal expression to bridge human creativity and computer-aided design, enabling a more seamless translation between conceptual ideas and visual form.

Inspired by this design scenario, we propose a designer-centered data augmentation paradigm by leveraging (1) a reference surface program prompt and (2) a design procedure prompt to synthesize CAD programs with more organic shapes. Our experiments demonstrate that, in comparison to the benchmark CAD program datasets, the CAD programs generated by our approach, after being interpreted to 3D CAD B-rep format, have improved diversity of shape complexity. In addition, the geometric properties (e.g., B-Spline ratio) of our generated data are more similar to those of industrial-grade CAD designs. Finally, our ablation study highlights the importance of the proposed reference surface prompt in driving the generation of B-Spline shapes and the advantage of program modularization for reflecting the proposed design procedures.


\section{Related Work}
CAD program generation \cite{wu2021iccv,wang2025texttocad,alam2025gencadimageconditionedcomputeraideddesign,rukhovich2025iccv,xu2024CADMLLM} has gained increased attention for generating 3D CAD designs. While parallel efforts such as primitive fitting \cite{liu2024sigraph, liu2024point2cad} and B-rep generation models \cite{xu2024brepgen,lee2025siggraph} are focused on 3D shape and topology representation, CAD program describes a CAD design as a sequence of design operations, and is more programmable and convertible to other CAD representation formats. 

Earlier works such as Fusion360 \cite{willis2020fusion}, DeepCAD \cite{wu2021iccv}, and GenCAD \cite{yu2025idetc}  leverage Domain Specific Languages (DSLs) (e.g., FeatureScript \cite{feasurescript}, structured text representation \cite{zhangflexcad}, CAD assembly JSON \cite{khan2024textcad}) to parse CAD operations and parameters. Based on a DSL, the generation of CAD programs can be performed by a deep reinforcement learning agent \cite{tang2020reinforcement} or by a generative model such as GAN \cite{gulrajani2017improved} and diffusion model \cite{ho2020denoising} in a latent space learned by a transformer \cite{NIPS2017_3f5ee243}. In addition, DSLs have been used to fine-tune LLMs for text-to-CAD generation \cite{wang2025texttocad,khan2024textcad} and VLMs for CAD generation conditioned by image and text \cite{alam2025gencadimageconditionedcomputeraideddesign,xu2024CADMLLM}. To improve the compatibility of DSL representation with LLMs, recent work \cite{li2025cvpr} has explored an augmented representation of CAD generation process which combines natural language-based text descriptions and parametric annotations in a hierarchical semantic description.

Python-based parametric CAD scripts such as CadQuery \cite{au_2024_10513848} and OpenSCAD \cite{OpenCascade} have been incorporated by many popular foundational LLMs, including OpenAI o3, Gemini, and Claude, into their code generation features. Taking advantage of this built-in feature, recent work such as CAD-recode \cite{rukhovich2025iccv} for point cloud-to-CAD generation and Cad-coder \cite{doris2025cad} for image-to-CAD generation has proposed fine-tuning open-source foundational LLM models such as Qwen \cite{qwen} and LLaVA \cite{liu2024improved} for conditional CAD generation. The rise of foundation LLM models in conditional CAD generation has created new opportunities for CAD data augmentation.

Despite the active research on CAD program generation mentioned above, current deep generative models (including LLMs) for CAD generation are still far from meeting industry's need for automated engineering design. Previous work \cite{kimmel2025position} has indicated most 3D CAD generation datasets such as DeepCAD \cite{wu2021iccv} and ABC \cite{Koch_2019_CVPR} have been intentionally
restricted in sketch-extrude operations that do not align with most real manufactured shapes. 
To narrow this gap and increase the shape complexity of publicly available CAD datasets, GenCAD \cite{alam2025gencadimageconditionedcomputeraideddesign} and CAD-MLLM \cite{xu2024CADMLLM} propose to develop multimodality datasets by using image prompts for data augmentation. However, the conditioning images used in those works are primarily rendered from open-source CAD program samples (i.e., DeepCAD), therefore constraining the shape distribution of the training dataset to shapes covered by existing CAD program datasets and preventing the deep learning models from accessing more sophisticated geometric features such as B-Spline geometries commonly seen in industry-level CAD objects. 

In comparison, our proposed approach ensures the generation of B-Spline geometries by including in our LLM prompts a Python script that defines a B-Spline-shaped reference surface. By combining this Python script with a natural language-based description of a design procedure, we command the CAD-generating LLMs to conform to the reference surface and sequentially influence other connected primitives, thus synthesizing CAD objects with B-Spline geometries. Our proposed design prompts which combines text description and reference surface CAD programs is compatible with any off-the-shape code-generating LLM and provides users with a conditioning mechanism for CAD generation that is both parameterizable and visualization.

\section{Proposed Method}
\label{sec:method}
\begin{figure}[!htbp]
    \centering
    \includegraphics[width=1\textwidth]{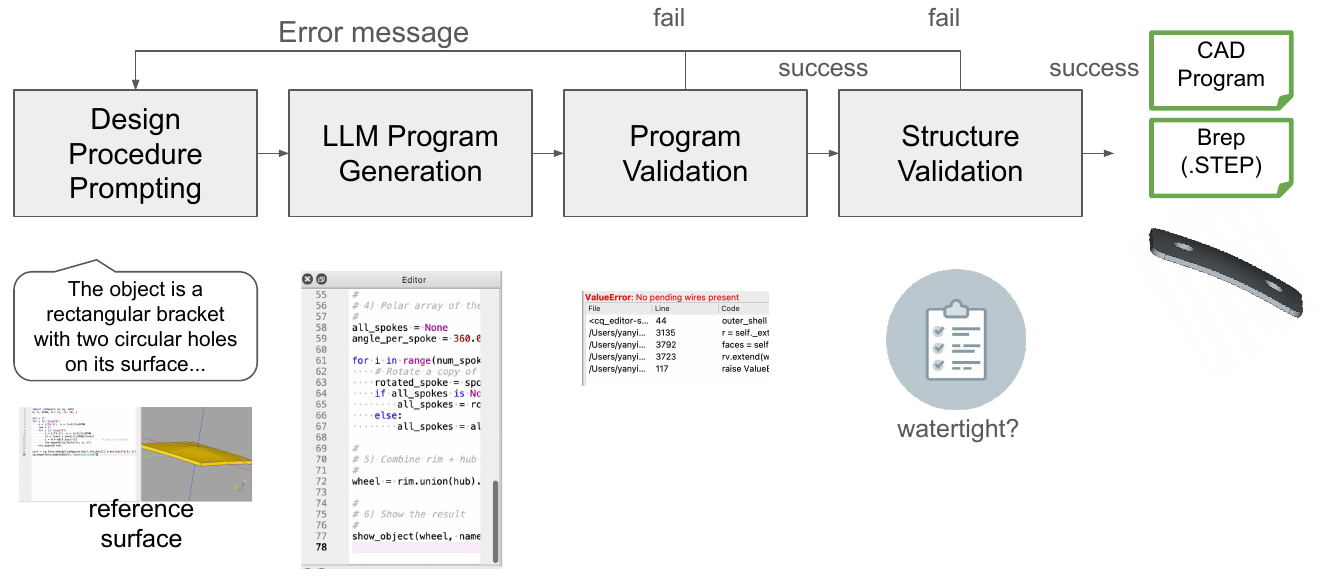}
    \caption{System overview: (1) Design procedure prompting takes a design description and a reference surface program as an input to formulate the design procedure. (2) The design prompt then condition an LLM to generate a CAD program. (3) Program validation executes the generated program to visualize a CAD Brep. (4) Structure validation checks the validity of the CAD Brep.}
    \label{fig:sys}
\end{figure}
The goal of this work is to develop a CAD program generation approach for data augmentation which enables the generation of complicated geometric properties commonly found in industrial practice. The core idea of our approach is a novel prompting strategy for LLM-based CAD program generation that is inspired by CAD model design practices from industry. 

\subsection{Design Procedure Prompting} \label{sec:method-prompt}
A formative study conducted with industrial designers to understand their design practices has introduced to us the following design procedure. The creation of a 3D CAD object starts from a given reference surface that the target CAD object is supposed to match. All subsequent operations that lead to the final CAD object will therefore influence the choice of the reference surface. In many industrial practices, the reference surface is chosen as a free-form B-spline shape for aesthetic or functional purposes. First, in exterior styling, B-spline organic curves are often used by designers to create smooth and flowing surfaces. Second, for interior components, ergonomic considerations are another key reason for using B-splines because these curves conform more naturally to a human body. In addition, from the perspective of structure, B-spline geometries are advantageous. In automotive body design, sharp creases or abrupt curvature transitions can lead to stress concentrations and starting points for buckling. A smooth, continuous B-spline surface helps distribute stress more evenly. Unlike a simple design procedure starting from basic sketch and extrusion (commonly found in public CAD program datasets), the B-Spline reference surface induces organic shaped primitives (e.g., faces and edges) parametrized by higher-order polynomial functions in the final design of the CAD object. The reference surface will be removed after the CAD object is created. 

To approximate this design procedure, we prompt an LLM CAD program generation model with a text description of the intended design and a reference surface input. The text prompt template is formatted as follows.

\fbox{text prompt := [prefix system prompt, design description, design context, postfix system prompt]}
\\\\\textbf{Prefix system prompt}: We instruct the program generation model to write a python-based CAD program as follows: \textit{``Use Python CadQuery library to write a CAD program of a bracket that is described as follows.''}
\\\textbf{Design description}: We provide the generation model with a description of an intended design, e.g., \textit{``The object is a rectangular bracket with two circular holes on its surface.''} as shown in Figure \ref{fig:sys}. 
\\\textbf{Design context}: we offer more detailed instructions on design requirement and procedure: \textit{''The shapes of the bracket look smooth. The bracket should conform to the curvature of the reference surface in the CAD program below. After the bracket is created, the reference surface should be removed.''} 
\\\textbf{Postfix system prompt}: We require that the generated CAD object is watertight solid and specify the path of the output STEP file after program execution. The details of postfix system prompt is reported in Appendix \ref{apx:postfix}. 

The content of both prefix and postfix system prompts remain the same for each instance of CAD generation while the design description varies to create diverse designs. The design context remains constant for the same target category of design (e.g., bracket) and is updated as a new target category (e.g., car wheel) is introduced. The example full prompts are reported in Appendix \ref{apx:full_prompt}

\subsection{Reference Surface Program}
Instead of using an image or a point cloud to represent a reference surface, we use a Python script representation of CAD program as an input prompt as shown in Figure \ref{fig:sys}. The reason for using a script-based program to represent a reference surface is that the script provides an accurate and expressive parametric description of 3D geometry that is well-understood by off-the-shelf LLMs extensively trained for Python code generation. In contrast, directly prompting a CAD program generating LLM with visual content tends to produce inaccurate geometry due to LLMs' limitation in learning cross-modality correlation, especially if the surface has a more organic shape instead of typical primitives.

To ensure a higher level of shape diversity in CAD data augmentation, we included four types for reference surfaces: Gaussian surface, saddle surface, wave surface and ripple surface - all of which are B-Spline surfaces. Each surface type is represented in a Python script using the CADquery library. To further increase shape diversity, we vary the parameters of each reference surface script to create more surface geometries; for example, we vary the curvature of a saddle surface from shallow to deep, with the variation automated by updating shape parameters in the Python script. In each instance of CAD generation, we select a reference surface and pair it with a design description to formulate a prompt. The design description data are provided in the experiment setting in Sec \ref{sec:exp-setting}.

\subsection{Program Generation and Validation}
As illustrated in Figure \ref{fig:sys}, the formatted prompt is fed to an LLM for CAD program generation. The LLM can be almost any popular off-the-shelf foundational LLM, and we chose to use OpenAI o3 \footnote{The snapshot o3-2025-04-16 at \url{https://platform.openai.com/docs/models/o3} was used. The parameter of reasoning effort was set to be high. The cost of the API was \$8 per 1M output tokens.} in our experiments. Each generated program goes through two agentic validation stages: (1) program validation to verify whether it can be converted to a boundary representation (B-rep) and exported to a STEP file and (2) structure validation to verify whether the generated B-rep is watertight and structurally feasible using the validity check proposed in DTGBrepGen \cite{li2025cvprdtgbrepgen}. 
Any identified errors in the CAD program are fed back to the prompt in an iterative process for self-correction. Once validated as a successful generation, the CAD program script is finalized and then converted to a B-rep.

\section{Experiments} \label{sec:experiment}

The purpose of the experiments is to evaluate whether the proposed data augmentation method can synthesize CAD designs that contain a larger portion of shape primitives with free-form organic shapes resembling actual designs for industry application. We first explain the evaluation metrics and existing datasets as the baselines, followed by a comparison to the geometric properties of the baselines and our generated CAD data. An ablation study is reported to investigate the effectiveness of different prompting strategies for generating organic shapes, and also illustrates how the proposed prompting strategies affect the programmed modules. Finally, we show the potential for extending the proposed approach to multiple design targets.

\subsection{Experiment Setting} \label{sec:exp-setting}
\textbf{Evaluation metrics}: The metrics include several geometric properties: (1) The number of lines of content in each STEP file. More lines of content in a STEP file generally indicates a more complex CAD design, as STEP files contain a formatted ASCII text description of geometric elements and their connectivity in a CAD object. (2) the number of faces and curves as defined in a Brep. (3) the proportion of B-rep STEP files with B-Spline faces and curves out of all generated STEP files. In addition, the B-Spline ratio $\beta_i$ of each CAD object $c_i$ is calculated as,
\begin{equation}    
\beta_i = [(f^b_i/f_i)+(e^b_i/e_i|)]/2, 
\end{equation}

where $f_i$ is the number of faces, $f^b_i$ is the number of B-Spline faces, $e_i$ is the number of curves and $e^b_i$ is the number of B-Spline curves. B-Spline, or more generally the non-uniform rational basis spline (NURBS), are commonly used in computer graphics and CAD design to represent a wide range of parameterized geometrical shapes, where the flexibility allows it to define more free-form shapes than standard primitives. Higher B-Spline Ratio means that more proportion of B-Spline shapes appear in the geometry. Note that these proxy metrics may not completely reflect true industrial-grade quality. Real-world quality also depends on factors like curvature continuity, fillet robustness, manufacturing tolerances, and feature intent, which are not assessed in the scope of this paper. 

\textbf{Baselines}:
In our experiments, the bracket is chosen as the target category for the generation of CAD, because brackets are a class of common mechanical parts and can be conveniently retrieved from the existing collection of commercial industry CAD designs (referred to as ``Industry'') and from the open-source CAD program dataset DeepCAD \cite{wu2021iccv} with keyword labels for filtering \cite{khan2024textcad} (referred as ``DeepCAD-b''). In addition to commercial industry designs and DeepCAD samples, we also included GenCAD dataset \footnote{downloaded from \url{https://github.com/ferdous-alam/GenCAD}} and CAD-MLLM dataset \footnote{downloaded from \url{https://huggingface.co/datasets/jingwei-xu-00/Omni-CAD/resolve/main/Omni-CAD.zip}} as benchmark baselines for a quantitative evaluation. Note that GenCAD and CAD-MLLM samples in our analysis includes CAD objects outside the bracket category as no keyword filtering is available. 
In addition, our approach requires design descriptions as part of the text prompt as mentioned in Sec \ref{sec:method}. For the experiments of bracket generation, the text descriptions \cite{khan2024textcad} associated with each of the DeepCAD bracket data are used as our design descriptions.

\subsection{Geometric Properties of Programmed CAD Data} 

\begin{table}
  \caption{A comparison of geometric properties with a commercial industry bracket dataset and baseline CAD program datasets DeepCAD-b, GenCAD and CAD-MLLM. (*) indicates that the data also include non-bracket objects.}
  \label{tbl:geomrtic-eval}
  \centering
  \begin{tabular}{llllll}
    \toprule
    \cmidrule(r){1-2}
    Bracket Data     & Industry     & DeepCAD-b & GenCAD* & CAD-MLLM* & Ours\\
    \midrule
    avg. \#lines (STEP) & 10099  & 1783 & 991  & 2402 & 4494 \\
    avg. \#faces     & 82.91 & 21.48  & 12.55  & 28.65 & 26.57 \\
    avg. \#curves     & 259.4       & 50.30 & 27.86 & 69.26 & 67.57 \\
    w/ B-Spline faces     & 100\%   & 0\%  & 0\%  & 0\% & 77\%\\
    w/ B-Spline curves     & 100\%   & 1\%  & 1\% & 1\% & 89\%\\
    B-Spline Ratio     & 0.5352       & 0.0004  & 0.0009 & 0.0008 & 0.2217 \\
    \bottomrule
  \end{tabular}
\end{table}

As reported in the first column from the right of Table \ref{tbl:geomrtic-eval}, our approach generates CAD programs where the corresponding B-rep data have a higher or similar average number of STEP lines, surfaces and curves in comparison to the existing CAD program datasets: DeepCAD-b, GenCAD and CAD-MLLM. This suggests that our approach is able to synthesize more designs with an organic shape without compromising geometric complexity. In particular, $77\%$ of the CAD objects generated by our approach contain B-Spline faces, whereas none of the CAD objects from DeepCAD-b, GenCAD and CAD-MLLM contain any B-Spline face. Similarly, $89\%$ of the CAD objects generated by our approach contain B-Spline curves, in contrast to only $1\%$ of the CAD objects from DeepCAD-b, GenCAD and CAD-MLLM containing B-Spline curves. In terms of the mean B-Spline Ratio in a CAD object, our generated CAD objects also outperform DeepCAD-b, GenCAD, and CAD-MLLM by a large margin, as shown in the bottom row of the table. This result is aligned with previous work \cite{kimmel2025position} that the existing public CAD datasets are much simpler than the commercial CAD data used in industry. It also demonstrates that our generated CAD objects have a more organic shape than the current open-source CAD program datasets, featuring a much closer resemblance to the CAD designs from industry where every sample contains B-Spline geometry and over $50\%$ of the shape primitives in an object are B-Splines.

\subsection{Diversity of Shape Complexity} 
\begin{figure}[!htbp]
    \centering
    \includegraphics[width=0.6\textwidth]{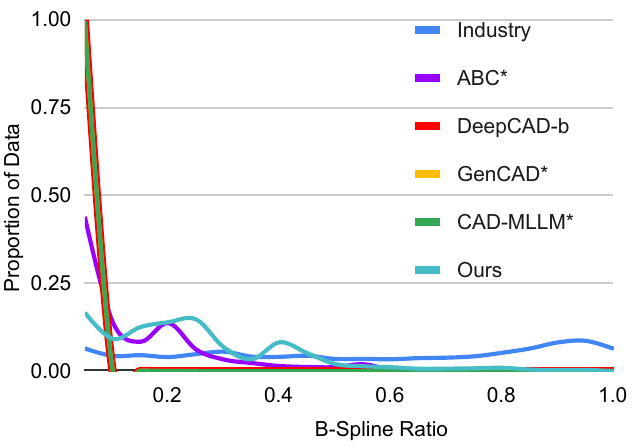}
    \caption{Distribution of B-Spline ratio over data samples: Industry, DeepCAD-b and ours only contain bracket data. ABC, GenCAD and CAD-MLLM include brackets and other objects.}
    \label{fig:bspline-distr}
\end{figure}
As shown in Figure \ref{fig:bspline-distr}, the CAD data generated by our approach are more evenly distributed over different B-Spline ratio, which suggests the capability to generate programmed CAD data that covers a wider range of shape complexity than DeepCAD-b, GenCAD and CAD-MLLM. Note that the distributions of DeepCAD-b, GenCAD, and CAD-MLLM over B-Spline ratio are very similar in a sense that a majority of the samples have zero B-Spline components. This is likely because GenCAD and CAD-MLLM are both built upon DeepCAD and therefore inherit the same limitation on the B-Spline-free operations adopted by DeepCAD. Compared with GenCAD and CAD-MLLM, our approach does not require image prompts and instead uses a script-based CAD program for reference surface to condition CAD program generation. Since both the script-based CAD program and the natural language-based text description are processed as text tokens by an LLM, no multimodal data processing of the LLM or VLM is required, which allows a more efficient and coherent generation of CAD designs and avoids the potential generation artifacts due to a suboptimal contrastive language-image pre-training. 

In addition to the existing CAD program datasets, we also include the ABC dataset\cite{Koch_2019_CVPR} (referred to as ABC-misc), by far the largest public CAD dataset. ABC-misc includes miscellaneous CAD STEP files without the corresponding CAD programs. Note that all publicly available baselines \ref{sec:exp-setting} are direct or indirect derivatives of ABC dataset by reversing a subset of the ABC 3D CAD objects to CAD programs using a limited set of CAD operations while discarding those that cannot be parsed successfully. ABC as a larger collection is considered to have more geometric diversity than the CAD program datasets. We conduct experiments on the commonly used test split of ABC \cite{liu2024sigraph,liu2024point2cad}. In comparison to the ABC dataset, our generated data is less skewed at low B-Spline region, offers a more significant representation in the higher B-Spline region, and is closer to the more evenly-distributed industrial bracket datasets (Industry).

\subsection{Ablation Study on Design Procedure Prompting}
\begin{figure}[!htbp]
    \centering
    \includegraphics[width=1\textwidth]{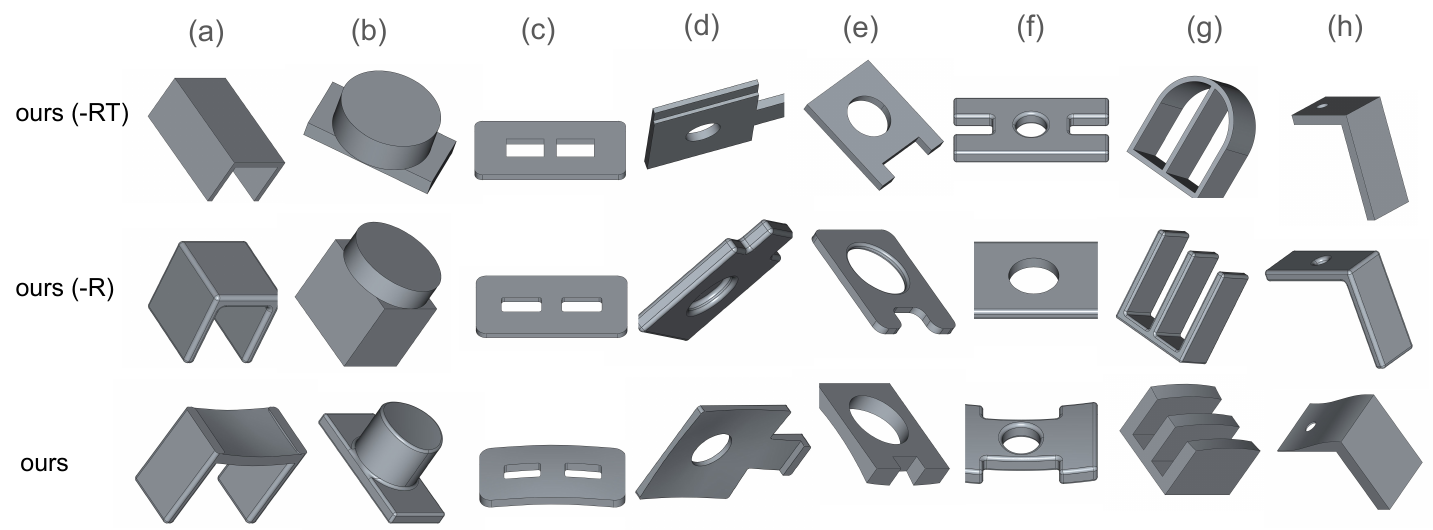}
    \caption{Examples of CAD B-rep visualization of our approach using reference surfaces, and the alternatives (-RT) and (-R) excluding reference surfaces. (-R) include a text guidance to prompt smooth and organic shapes. Ours with reference surfaces can generate more B-Spline shapes.}
    \label{fig:qual-comparison}
\end{figure}
\begin{table} [!htbp]
  \caption{An ablation study of our approach, testing the alternative prompts (-RT) and (-R) without using reference surfaces.}
  \label{tbl:ablation}
  \centering
  \begin{tabular}{llll}
    \toprule
    \cmidrule(r){1-2}
    Approach     & Ours(-RT)     & Ours(-R) & Ours \\
    \midrule
    avg. \#lines (STEP) & 1225 & 2992& 4494 \\
    avg. \#faces & 14.86 & 34.56& 26.57 \\
    avg. \#curves & 35.17 & 75.89 & 67.57 \\
    w/ B-Spline faces & 2\%& 18\%& 77\% \\
    w/ B-Spline curves & 6\%& 27\%& 89\% \\
    B-Spline Ratio & 0.0085 & 0.0478& 0.2217 \\
    \bottomrule
  \end{tabular}
\end{table}
An ablation study is conducted to test the effectiveness of using reference surface prompts and the alternatives, referred to as ``ours(-RT)'' and ``ours(-R)''. Ours(-RT) completely removes the design context (instructing the design procedure) in the text prompt (Sec. \ref{sec:method-prompt}) and excludes the input reference surface program. Ours(-R) is similar to ours(-RT), but replaces the original design context with a text-based shape guidance, i.e., ``The shapes of the bracket look smooth and organic.'' This is to test to what extent a simple text-based shape guidance may affect the generated results. 

As shown in Table \ref{tbl:ablation}, the B-Spline ratio is largely decreased if the reference surface prompt or the design procedure prompt are not used, i.e., ours vs. ours(-RT), suggesting the importance of the proposed design procedure prompting. Ours(-R), by excluding the reference surface prompt and using text-based shape guidance, only slightly improves the B-Spline ratio, i.e., ours(-R) vs. ours(-RT). This suggests surface program is the key factor to guide the program generation to yield more B-Splines.

Interestingly, the average number of faces and curves of ours(-R) are higher than ours. We suspect that the LLMs program generation model attempts to generate more standard primitives to illustrate the text guidance regarding smooth and organic shapes, instead of using B-Splines to concisely represent the intended shapes. This suggests the limitation of natural language guidance for complicated shapes.

Figure \ref{fig:qual-comparison} presents the generated CAD B-rep examples. The same design description is applied to each design in a column. Ours has a more variety of curvatures than ours(-RT) and ours(-R), attributed to the intent to match varied reference surfaces.

\subsection{Program Modularization for Design Procedure}
\begin{figure}[!htbp]
    \centering
    \includegraphics[width=1\textwidth]{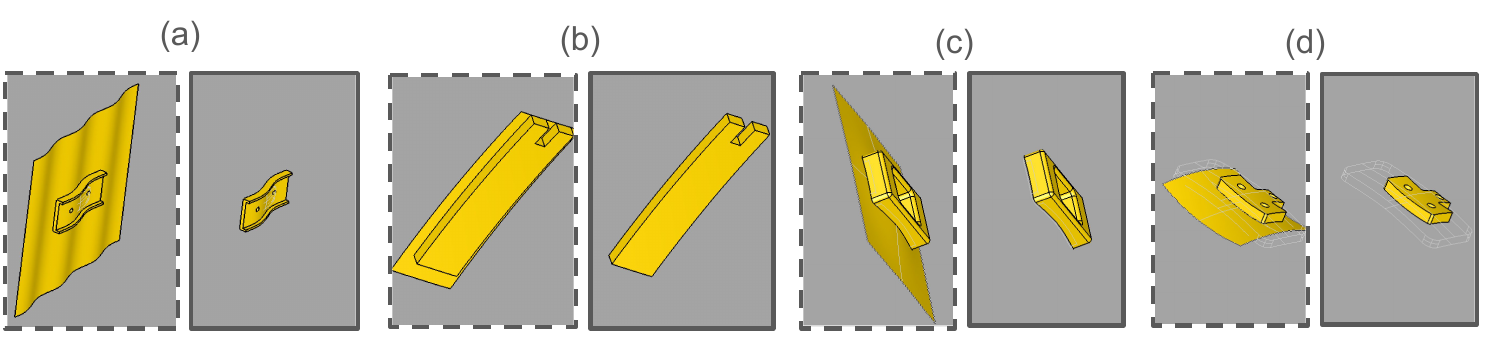}
    \caption{Visualization of the generated program modules. The left side of each column is a generated bracket with its reference surface, and the right side is the bracket after the reference surface is removed.}
    \label{fig:ref-surface-results}
\end{figure}
The design procedure conditioned on a reference surface is visualized in Figure \ref{fig:ref-surface-results}, where CQ-Editor \footnote{https://github.com/CadQuery/CQ-editor} is used to compare the programmed modules with (left) and without (right) the presence of the reference surface. Each created bracket naturally comprises more B-Splines as a result of matching the given reference surface. The variation of reference surfaces creates the necessary shape diversity for data augmentation. Unlike voxel-based deformation of 3D objects, the deformation by our approach is implemented at the CAD program level using LLM prompts, which provides a smoother and more precise shape control for procedural generation of 3D shapes.

\subsection{Generalizability to other Design Targets}
\begin{figure}
    \centering
    \includegraphics[width=1\textwidth]{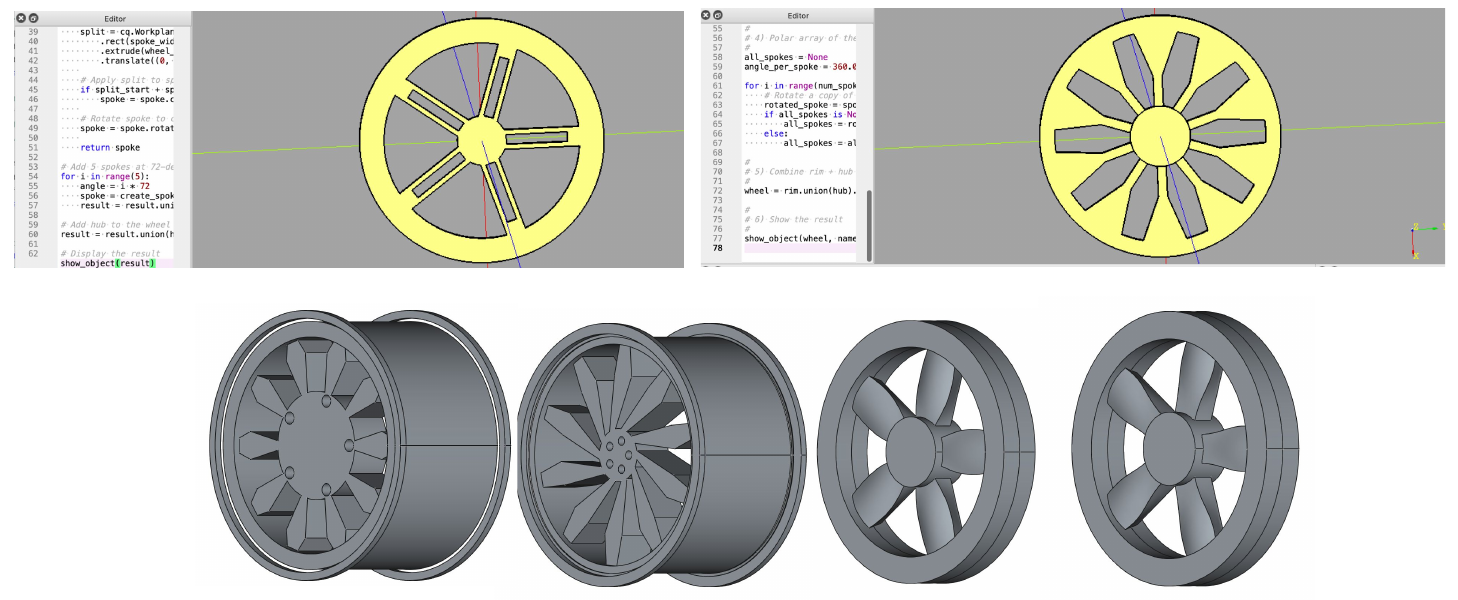}
    \caption{Examples of generated wheels. The top row are the program visualization. The bottom row are the CAD Breps. Reference surfaces are used to generate the two wheels in the bottom right. }
    \label{fig:wheel-vis}
\end{figure}
To test the generalizability to more design targets, a qualitative experiment is conducted on car wheels using our approach. We leverage the design patterns used in the prior paper \cite{wang2024idetc} to formulate the design descriptions. The upper plots in Figure \ref{fig:wheel-vis} shows the generated program and the visualization in CQ-Editor, where the design descriptions specify the number of spokes (5 vs. 10) and the type of spokes (double-spoke vs. Y-spoke). The lower plots shows a visualization of the STEP files of more generated wheels, prompted by the design description with regard to the hub and the barrel. A reference surface is used to prompt the spoke shapes of the two wheels counting from right to left.

\section{Limitation and Discussion}
\begin{table}
  \caption{The proportion of generated data that requires more than 5 iterations of re-generation due to program execution errors or structure validation failures.}
  \label{tbl:regeneration}
  \centering
  \begin{tabular}{llll}
    \toprule
    \cmidrule(r){1-2}
    Approach     & Ours(-RT)     & Ours(-R) & Ours \\
    \midrule
    require > 5 iterations & 12\%  & 18\% & 21\%\\
    \bottomrule
  \end{tabular}
\end{table}
While design procedure is demonstrated to be useful for CAD program generation, it also creates more complicated programs that may result in more execution errors and invalid structures. This highlights the importance of the program validation and the structure validation used in our proposed system (Fig. \ref{fig:sys}). Validation feedback and iterative generation largely improve validity, but also increase the iterations needed for a successful generation, as reported in Table \ref{tbl:regeneration}.

Our approach still requires the programs of reference surfaces. While we can sample a variety of reference shapes for data augmentation purpose, script-based surface programs are less available for prompting precision control generation. Logging of operations is required when a designer is creating a reference surface; without logged operations, a reverse engineering process would be required to obtain/generate a reference surface program. Nevertheless, reference surface program generation is a reduced task of general CAD program generation and can be critical to progressively address the challenges.

Our approach cannot precisely control where the bracket is supposed to match the reference surface. In Figure \ref{fig:qual-comparison} (a) a reference surface is applied to the base, while in (g) a reference surface is applied to the three feet. Precision control requires a deeper understanding for CAD parameterization, which might be assisted by a semi-automated way (e.g., human intervention) or LLM models finetuned by pairs of precise input and output. We hope our data augmentation work can contribute to scaling up diverse data for finetuning precision generation models.   
\section{Conclusion}
In the presence of the current challenge of lacking CAD program datasets with complicated geometry, an effective data augmentation strategy to enrich CAD program datasets could be an essential piece to scaling up the deep learning models' performance in CAD program generation. Inspired by CAD design practice from industry, we propose a novel procedure for LLM prompting which uses a reference surface program to guide the generation of CAD programs. Experiment results demonstrate that the generated CAD programs by our proposed approach has shown a significant improvement on the diversity of shape complexity compared with baseline CAD programs for LLMs. A comparison with practical CAD design from industry indicates that our method is able to narrow the gap in geometric complexity between synthesized CAD designs from machine learning and practical CAD designs used in industry applications.

\bibliographystyle{plain}
\bibliography{references}

@InProceedings{wu2021iccv,
    author    = {Wu, Rundi and Xiao, Chang and Zheng, Changxi},
    title     = {DeepCAD: A Deep Generative Network for Computer-Aided Design Models},
    booktitle = {Proceedings of the IEEE/CVF International Conference on Computer Vision (ICCV)},
    month     = {October},
    year      = {2021},
    pages     = {6772-6782}
}

@article{willis2020fusion,
    title={Fusion 360 Gallery: A Dataset and Environment for Programmatic CAD Construction from Human Design Sequences},
    author={Karl D. D. Willis and Yewen Pu and Jieliang Luo and Hang Chu and Tao Du and Joseph G. Lambourne and Armando Solar-Lezama and Wojciech Matusik},
    journal={ACM Transactions on Graphics (TOG)},
    volume={40},
    number={4},
    year={2021},
    publisher={ACM New York, NY, USA}
}

@misc{feasurescript,
    author = {Onshape featurescript.},
    title = {\url{https://cad.onshape.com/
FsDoc/}}
}

@inproceedings{
wang2025texttocad,
title={Text-to-{CAD} Generation Through Infusing Visual Feedback in Large Language Models},
author={Ruiyu Wang and Yu Yuan and Shizhao Sun and Jiang Bian},
booktitle={Forty-second International Conference on Machine Learning (ICML)},
year={2025},
}

@inproceedings{
li2025cvpr,
title={CAD-Llama: Leveraging Large Language Models for Computer-Aided Design Parametric 3D Model Generation},
author={Jiahao Li and Weijian Ma and Xueyang Li and Yunzhong Lou and Guichun Zhou and Xiangdong Zhou},
booktitle={Proceedings of the IEEE/CVF Computer Vision and Pattern Recognition Conference (CVPR)},
year={2025},
}

@Inproceedings{khan2024textcad,
title={Text2CAD: Generating Sequential {CAD} Designs from Beginner-to-Expert Level Text Prompts},
author={Mohammad Sadil Khan and Sankalp Sinha and Sheikh Talha Uddin and Didier Stricker and Sk Aziz Ali and Muhammad Zeshan Afzal},
booktitle = {Advances in Neural Information Processing Systems},
pages = {7552--7579},
publisher = {Curran Associates, Inc.},
year={2024},
volume = {37},
url = {https://proceedings.neurips.cc/paper_files/paper/2024/file/0e5b96f97c1813bb75f6c28532c2ecc7-Paper-Conference.pdf},
}

@misc{xu2024CADMLLM,
      title={CAD-MLLM: Unifying Multimodality-Conditioned CAD Generation With MLLM}, 
      author={Jingwei Xu and Chenyu Wang and Zibo Zhao and Wen Liu and Yi Ma and Shenghua Gao},
      year={2024},
      eprint={2411.04954},
      archivePrefix={arXiv},
      primaryClass={cs.CV}
}

@misc{au_2024_10513848,
  author       = {AU and
                  Jeremy Wright and
                  thebluedirt and
                  Marcus Boyd and
                  Lorenz and
                  Innovations Technology Solutions and
                  Hasan Yavuz ÖZDERYA and
                  Bruno Agostini and
                  Jojain and
                  Michael Greminger and
                  Seth Fischer and
                  Justin Buchanan and
                  cactrot and
                  huskier and
                  Ruben and
                  iulianOnofrei (U-lee-aan) and
                  Miguel Sánchez de León Peque and
                  Martin Budden and
                  Hecatron and
                  Peter Boin and
                  Wink Saville and
                  Pavel M. Penev and
                  Bryan Weissinger and
                  M. Greyson Christoforo and
                  Jack Case and
                  AGD and
                  Paul Jurczak and
                  nopria and
                  moeb and
                  jdegenstein},
  title        = {CadQuery/cadquery: CadQuery 2.4.0},
  month        = {jan},
  year         = {2024},
  publisher    = {Zenodo},
  version      = {2.4.0},
  doi          = {10.5281/zenodo.10513848},
  url          = {https://doi.org/10.5281/zenodo.10513848},
}

@InProceedings{rukhovich2025iccv,
    author    = {Danila Rukhovich and Elona Dupont and Dimitrios Mallis and Kseniya Cherenkova and Anis Kacem and Djamila Aouada},
    title     = {CAD-Recode: Reverse Engineering CAD Code from Point Clouds},
    booktitle = {Proceedings of the IEEE/CVF International Conference on Computer Vision (ICCV)},
    month     = {October},
    year      = {2025}
}

@misc{alam2025gencadimageconditionedcomputeraideddesign,
      title={GenCAD: Image-Conditioned Computer-Aided Design Generation with Transformer-Based Contrastive Representation and Diffusion Priors}, 
      author={Md Ferdous Alam and Faez Ahmed},
      year={2025},
      eprint={2409.16294},
      archivePrefix={arXiv},
      primaryClass={cs.CV},
      url={https://arxiv.org/abs/2409.16294}, 
}

@article{yu2025idetc,
    author = {Yu, Nomi and Alam, Md Ferdous and Hart, A. John and Ahmed, Faez},
    title = {GenCAD-3D: CAD Program Generation using Multimodal Latent Space Alignment and Synthetic Dataset Balancing},
    journal = {Journal of Mechanical Design},
    pages = {1-17},
    year = {2025},
    issn = {1050-0472},
    doi = {10.1115/1.4069276},
    url = {https://doi.org/10.1115/1.4069276},
    eprint = {https://asmedigitalcollection.asme.org/mechanicaldesign/article-pdf/doi/10.1115/1.4069276/7524252/md-25-1380.pdf},
}

@article{doris2025cad,
  title={CAD-Coder: An Open-Source Vision-Language Model for Computer-Aided Design Code Generation},
  author={Doris, Anna C and Alam, Md Ferdous and Nobari, Amin Heyrani and Ahmed, Faez},
  journal={arXiv preprint arXiv:2505.14646},
  year={2025}
}

@article{qwen,
  title={Qwen Technical Report},
  author={Jinze Bai and Shuai Bai and Yunfei Chu and Zeyu Cui and Kai Dang and Xiaodong Deng and Yang Fan and Wenbin Ge and Yu Han and Fei Huang and Binyuan Hui and Luo Ji and Mei Li and Junyang Lin and Runji Lin and Dayiheng Liu and Gao Liu and Chengqiang Lu and Keming Lu and Jianxin Ma and Rui Men and Xingzhang Ren and Xuancheng Ren and Chuanqi Tan and Sinan Tan and Jianhong Tu and Peng Wang and Shijie Wang and Wei Wang and Shengguang Wu and Benfeng Xu and Jin Xu and An Yang and Hao Yang and Jian Yang and Shusheng Yang and Yang Yao and Bowen Yu and Hongyi Yuan and Zheng Yuan and Jianwei Zhang and Xingxuan Zhang and Yichang Zhang and Zhenru Zhang and Chang Zhou and Jingren Zhou and Xiaohuan Zhou and Tianhang Zhu},
  journal={arXiv preprint arXiv:2309.16609},
  year={2023}
}

@inproceedings{NIPS2017_3f5ee243,
 author = {Vaswani, Ashish and Shazeer, Noam and Parmar, Niki and Uszkoreit, Jakob and Jones, Llion and Gomez, Aidan N and Kaiser, \L ukasz and Polosukhin, Illia},
 booktitle = {Advances in Neural Information Processing Systems},
 editor = {I. Guyon and U. Von Luxburg and S. Bengio and H. Wallach and R. Fergus and S. Vishwanathan and R. Garnett},
 pages = {},
 publisher = {Curran Associates, Inc.},
 title = {Attention is All you Need},
 url = {https://proceedings.neurips.cc/paper_files/paper/2017/file/3f5ee243547dee91fbd053c1c4a845aa-Paper.pdf},
 volume = {30},
 year = {2017}
}

@inproceedings{
kimmel2025position,
title={Position: You Can't Manufacture a Ne{RF}},
author={MA Kimmel and Mueed Ur Rehman and Yonatan Bisk and Gary K. Fedder},
booktitle={Forty-second International Conference on Machine Learning Position Paper Track},
year={2025},
url={https://openreview.net/forum?id=kJzB6lQmcb}
}

@InProceedings{Koch_2019_CVPR,
author = {Koch, Sebastian and Matveev, Albert and Jiang, Zhongshi and Williams, Francis and Artemov, Alexey and Burnaev, Evgeny and Alexa, Marc and Zorin, Denis and Panozzo, Daniele},
title = {ABC: A Big CAD Model Dataset For Geometric Deep Learning},
booktitle = {The IEEE Conference on Computer Vision and Pattern Recognition (CVPR)},
month = {June},
year = {2019}
}

@article{liu2024sigraph,
author = {Liu, Yilin and Chen, Jiale and Pan, Shanshan and Cohen-Or, Daniel and Zhang, Hao and Huang, Hui},
title = {Split-and-Fit: Learning B-Reps via Structure-Aware Voronoi Partitioning},
year = {2024},
issue_date = {July 2024},
publisher = {Association for Computing Machinery},
address = {New York, NY, USA},
volume = {43},
number = {4},
issn = {0730-0301},
url = {https://doi.org/10.1145/3658155},
doi = {10.1145/3658155},
journal = {ACM Trans. Graph.},
month = jul,
articleno = {108},
numpages = {13}
}

@inproceedings{liu2024point2cad,
  title={Point2CAD: Reverse Engineering CAD Models from 3D Point Clouds},
  author={Liu, Yujia and Obukhov, Anton and Wegner, Jan Dirk and Schindler, Konrad},
  booktitle={Proceedings of the IEEE/CVF Conference on Computer Vision and Pattern Recognition},
  pages={3763--3772},
  year={2024}
}

@article{xu2024brepgen,
		  title={Brepgen: A b-rep generative diffusion model with structured latent geometry},
		  author={Xu, Xiang and Lambourne, Joseph and Jayaraman, Pradeep and Wang, Zhengqing and Willis, Karl and Furukawa, Yasutaka},
		  journal={ACM Transactions on Graphics (TOG)},
		  volume={43},
		  number={4},
		  pages={1--14},
		  year={2024},
		  publisher={ACM New York, NY, USA}
		}

@inproceedings{lee2025siggraph,
author = {Lee, Mingi and Zhang, Dongsu and Jambon, Cl\'{e}ment and Kim, Young Min},
title = {BrepDiff: Single-Stage B-rep Diffusion Model},
year = {2025},
isbn = {9798400715402},
publisher = {Association for Computing Machinery},
address = {New York, NY, USA},
url = {https://doi.org/10.1145/3721238.3730698},
doi = {10.1145/3721238.3730698},
articleno = {103},
numpages = {11},
location = {
},
booktitle = {SIGGRAPH Conference Papers '25}
}

@inproceedings{
pfaff2021learning,
title={Learning Mesh-Based Simulation with Graph Networks},
author={Tobias Pfaff and Meire Fortunato and Alvaro Sanchez-Gonzalez and Peter Battaglia},
booktitle={International Conference on Learning Representations},
year={2021},
url={https://openreview.net/forum?id=roNqYL0_XP}
}

@article{li2022shapeoptimization,
title = {Machine learning in aerodynamic shape optimization},
journal = {Progress in Aerospace Sciences},
volume = {134},
pages = {100849},
year = {2022},
issn = {0376-0421},
doi = {https://doi.org/10.1016/j.paerosci.2022.100849},
url = {https://www.sciencedirect.com/science/article/pii/S0376042122000410},
author = {Jichao Li and Xiaosong Du and Joaquim R.R.A. Martins}
}

@inproceedings{wang2024idetc,
    author = {Wang, Ye and Damen, Nicole B. and Gale, Thomas and Seo, Voho and Shayani, Hooman},
    title = {Inspired by AI? A Novel Generative AI System to Assist Conceptual Automotive Design},
    volume = {Volume 6: 36th International Conference on Design Theory and Methodology (DTM)},
    booktitle = {International Design Engineering Technical Conferences and Computers and Information in Engineering Conference},
    year = {2024},
    month = {08}
}

@inproceedings{li2025cvprdtgbrepgen,
  title={DTGBrepGen: A Novel B-rep Generative Model through Decoupling Topology and Geometry},
  author={Jing Li and Yihang Fu and Falai Chen},
  booktitle={Proceedings of the IEEE/CVF Conference on Computer Vision and Pattern Recognition},
  year={2025}
}

@article{gulrajani2017improved,
  title={Improved training of wasserstein gans},
  author={Gulrajani, Ishaan and Ahmed, Faruk and Arjovsky, Martin and Dumoulin, Vincent and Courville, Aaron C},
  journal={Advances in neural information processing systems},
  volume={30},
  year={2017}
}

@article{ho2020denoising,
  title={Denoising diffusion probabilistic models},
  author={Ho, Jonathan and Jain, Ajay and Abbeel, Pieter},
  journal={Advances in neural information processing systems},
  volume={33},
  pages={6840--6851},
  year={2020}
}

@inproceedings{tang2020reinforcement,
  title={Reinforcement learning for integer programming: Learning to cut},
  author={Tang, Yunhao and Agrawal, Shipra and Faenza, Yuri},
  booktitle={International conference on machine learning},
  pages={9367--9376},
  year={2020},
  organization={PMLR}
}

@inproceedings{zhangflexcad,
  title={FlexCAD: Unified and Versatile Controllable CAD Generation with Fine-tuned Large Language Models},
  author={Zhang, Zhanwei and Sun, Shizhao and Wang, Wenxiao and Cai, Deng and Bian, Jiang},
  booktitle={The Thirteenth International Conference on Learning Representations},
year={2024}
}

@misc{OpenCascade,
  author = "{Open\,Cascade S.A.S.}",
  title = "{Open Cascade Technology (OCCT)}",
  year = {1999},
  url = {https://www.opencascade.com/},
  note = {Open‑source platform for 3D CAD/CAM/CAE; LGPL‑2.1 with exception},
}

@inproceedings{liu2024improved,
  title={Improved baselines with visual instruction tuning},
  author={Liu, Haotian and Li, Chunyuan and Li, Yuheng and Lee, Yong Jae},
  booktitle={Proceedings of the IEEE/CVF conference on computer vision and pattern recognition},
  pages={26296--26306},
  year={2024}
}








\appendix
\section{Postfix System Prompt} \label{apx:postfix}
The details of postfix system prompt is reported as below.

``Make sure the generated CAD model is watertight solid. Please export the generated CAD model to output.stl file and output.step file. Please do not visualize it. Here is the document of CadQuery for your reference (\url{https://cadquery.readthedocs.io/en/latest/index.html}). Do not output explanation.''

\section{Example Full Prompts} \label{apx:full_prompt}
The prefix system prompt, the design description, the design context and the postfix system prompt are color coded in \textcolor{gray}{gray}, \textcolor{blue}{blue}, \textcolor{purple}{purple} and \textcolor{brown}{brown}, repspectively, followed by the program of a reference surface.

Example 1:\\
\textcolor{gray}{Use Python CadQuery library to write a CAD program of a bracket that is described as follows.} \textcolor{blue}{The object is a U-shaped, open-ended, three-dimensional structure with a flat bottom and curved edges. It resembles a bracket or support structure.} \textcolor{purple}{The shapes of the bracket look smooth. The bracket should conform to the curvature of the reference surface in the CAD program below. After the bracket is created, the reference surface should be removed.} \textcolor{brown}{Make sure the generated CAD model is watertight solid. Please export the generated CAD model to output.stl file and output.step file. Please do not visualize it. Here is the document of CadQuery for your reference (https://cadquery.readthedocs.io/en/latest/index.html). Do not output explanation.}

\begin{lstlisting}[language=Python]
# saddle.py
import cadquery as cq, math
U, V, SPAN, CURV = 300, 300, 50, 0.004

net = []
for i in range(U):
    u = i/(U-1);  x = (u-0.5)*SPAN
    row = []
    for j in range(V):
        v = j/(V-1);  y = (v-0.5)*SPAN
        z = CURV*(x**2 - y**2)        
        row.append(cq.Vector(x, y, z))
    net.append(row)

surf = cq.Face.makeSplineApprox(net)
cq.exporters.export(surf, "saddle.step")
\end{lstlisting}

Example 2:\\
\textcolor{gray}{Use Python CadQuery library to write a CAD program of a bracket that is described as follows. }\textcolor{blue}{A rectangular bracket with two holes and two slots. }\textcolor{purple}{The shapes of the bracket look smooth. The bracket should conform to the curvature of the reference surface in the CAD program below. After the bracket is created, the reference surface should be removed. }\textcolor{brown}{Make sure the generated CAD model is watertight solid. Please export the generated CAD model to output.stl file and output.step file. Please do not visualize it. Here is the document of CadQuery for your reference (https://cadquery.readthedocs.io/en/latest/index.html). Do not output explanation. }
\begin{lstlisting}[language=Python]
import cadquery as cq, math
U, V, SPAN, H = 100, 100, 100, 7

net = []
for i in range(U):
    u = i/(U-1);  x = (u-0.5)*SPAN
    row = []
    for j in range(V):
        v = j/(V-1);  y = (v-0.5)*SPAN
        r2 = (x**2 + y**2)/((SPAN/3)**2)
        z = H * math.exp(-r2)                # Gaussian height
        row.append(cq.Vector(x, y, z))
    net.append(row)

surf = cq.Face.makeSplineApprox(net).thicken(2).translate((0,0,-1))
cq.exporters.export(surf, "gaussian.step")
\end{lstlisting}

\end{document}